\pdfoutput=1
\documentclass[11pt,a4paper]{article}

\usepackage[utf8]{inputenc}
\usepackage[T1]{fontenc}
\usepackage[margin=1in]{geometry}
\usepackage{graphicx}
\usepackage{amsmath,amssymb}
\usepackage{booktabs}
\usepackage{tabularx}
\usepackage{enumitem}
\usepackage{xcolor}
\usepackage{float}
\usepackage{caption}
\usepackage{subcaption}
\usepackage{fancyhdr}
\usepackage{titlesec}
\usepackage{abstract}
\usepackage{authblk}
\usepackage{algorithm}
\usepackage{algpseudocode}
\usepackage{listings}
\usepackage{natbib}
\usepackage{tikz}
\usetikzlibrary{shapes.geometric, arrows.meta, positioning, fit, backgrounds}
\usepackage{hyperref}
\usetikzlibrary{shapes.geometric, arrows.meta, positioning, fit, backgrounds}

\definecolor{linkblue}{HTML}{2E86C1}
\definecolor{darkblue}{HTML}{1B4F72}
\definecolor{codebg}{HTML}{F5F6FA}
\definecolor{codeframe}{HTML}{D5D8DC}

\hypersetup{
    colorlinks=true,
    linkcolor=darkblue,
    citecolor=darkblue,
    urlcolor=linkblue,
    bookmarksdepth=3
}

\title{
    \vspace{-1cm}
    {\LARGE\textbf{Machine Learning as a Tool (MLAT):}} \\[0.4cm]
    {\large A Framework for Integrating Statistical ML Models \\
    as Callable Tools within LLM Agent Workflows}
}

\author[1]{Edwin Chen\thanks{Corresponding author: \href{mailto:edwin@legacyai.agency}{edwin@legacyai.agency}}}
\author[1]{Zulekha Bibi}
\affil[1]{Legacy AI LLC}

\date{
    February 2026 \\[0.3cm]
    \small Submitted to the Gemini 3 Hackathon \\
    \small \url{https://devpost.com/software/pitchcraft}
}

\begin{document}

\maketitle
\thispagestyle{fancy}

\begin{abstract}
\noindent
We introduce \textbf{Machine Learning as a Tool (MLAT)}, a design pattern in which pre-trained statistical ML models are exposed as callable tools within LLM agent workflows, enabling the orchestrating agent to invoke real-time predictions and reason about their outputs contextually. Unlike conventional pipelines that treat ML inference as a static preprocessing step, MLAT positions the ML model as a first-class tool alongside web search, database queries, and API calls, allowing the LLM to decide \textit{when} and \textit{how} to invoke the model based on conversational context. Despite the naturalness of this pattern, it appears to be underexplored in both the academic literature on agentic AI and in production system architectures.

To validate MLAT, we present \textbf{PitchCraft}, a pilot production system that transforms discovery call recordings into professional proposals with ML-predicted pricing. PitchCraft implements MLAT through a single LLM workflow containing two Gemini-powered agents: a \textit{Research Agent} that performs prospect intelligence gathering via parallel tool calls, and a \textit{Draft Agent} that invokes an XGBoost pricing model as a tool call, reasons about the prediction, and generates a complete proposal via structured output parsing. The XGBoost model, trained on 70 examples (40 real agency deals augmented with 30 human-verified synthetic records), achieves $R^2 = 0.807$ on held-out test data with MAE of \$3,688. The complete system reduces proposal generation from 3+ hours to under 10 minutes.

We detail the MLAT framework formally, the structured output parsing architecture using Gemini's JSON schema capabilities, the ML methodology under extreme data scarcity (7:1 sample-to-feature ratio), group-aware cross-validation to prevent data leakage, and a sensitivity analysis demonstrating that the model has learned economically meaningful feature relationships. We argue that MLAT's applicability extends to any domain requiring quantitative estimation combined with contextual reasoning.

\vspace{0.3cm}
\end{abstract}

\vspace{0.5cm}

\section{Introduction}

\subsection{Motivation: The Missing Tool in Agentic AI}

Modern LLM agent frameworks support \textit{tool-calling}---the orchestrating LLM can invoke external functions such as web search, database queries, code execution, or API calls during its reasoning process. Frameworks such as LangChain~\citep{langchain2023}, AutoGen~\citep{wu2023autogen}, CrewAI, and n8n~\citep{n8n2025} have formalized multi-agent orchestration patterns where specialized agents coordinate through shared tool registries.

However, the tools available to these agents are overwhelmingly \textit{API-based services}. The use of pre-trained statistical ML models---XGBoost, random forests, logistic regression, neural networks trained on domain-specific tabular data---as agent-callable tools appears to be uncommon in both published literature and production system descriptions. This is surprising: for many business-critical tasks (pricing, risk assessment, demand forecasting, resource estimation), statistical ML models trained on historical data outperform LLMs at quantitative prediction, while LLMs excel at the contextual reasoning and natural language generation that surrounds those predictions.

We formalize this gap as the \textbf{Machine Learning as a Tool (MLAT)} pattern and argue it represents a natural but underexplored extension of tool-calling in agentic AI.

\subsection{The Proposal Bottleneck}

Quote-based service businesses---AI agencies, construction firms, consulting practices, home service providers---share a universal challenge: the speed at which they generate accurate, personalized proposals directly determines close rate. Research on lead response time consistently shows that faster follow-up correlates strongly with higher conversion~\citep{james2007lead}, yet most agencies spend 3--5 hours per proposal through a manual pipeline of transcript review, prospect research, scope estimation, and pricing calculation.

\subsection{The MLAT Hypothesis}

We hypothesize that exposing a trained ML model as a tool call---rather than embedding it as a static pipeline stage---provides three distinct advantages:

\begin{enumerate}[leftmargin=2em]
    \item \textbf{Contextual invocation:} The LLM agent decides \textit{when} to invoke the ML model based on conversational context, rather than invoking it unconditionally.
    \item \textbf{Interpretive reasoning:} The agent reasons about the prediction, explains rationale, and adjusts recommendations based on qualitative factors the ML model cannot capture.
    \item \textbf{Compositional flexibility:} The ML model becomes interchangeable and upgradeable without altering the agent's workflow logic, following the same pattern as other tool integrations.
\end{enumerate}

\subsection{Contributions}

This paper makes the following contributions:

\begin{itemize}[leftmargin=2em]
    \item We formalize \textbf{Machine Learning as a Tool (MLAT)} as a design pattern, including a formal definition and comparison with conventional ML integration patterns.
    \item We detail an \textbf{agent architecture} implementing MLAT with Gemini's structured output parsing, showing how JSON schemas bridge LLM reasoning and ML feature vectors.
    \item We present a methodology for training \textbf{XGBoost regression under extreme data scarcity} ($N=70$) with group-aware cross-validation and human-verified synthetic data augmentation.
    \item We validate MLAT through \textbf{PitchCraft}, a pilot production system deployed at Legacy AI LLC, reporting real-world impact and a sensitivity analysis confirming economically meaningful learned relationships.
\end{itemize}

\section{Related Work}

\subsection{LLM Tool-Calling and Agentic Frameworks}

The tool-calling paradigm, popularized by OpenAI's function calling~\citep{openai2023function} and adopted by Google's Gemini~\citep{gemini2024}, Anthropic's Claude, and open-source frameworks, enables LLMs to invoke external functions during generation. Schick et al.~\citep{schick2023toolformer} demonstrated that language models can learn to use tools autonomously, while Yao et al.~\citep{yao2023react} formalized the ReAct pattern of interleaving reasoning and action.

Despite this progress, the tools available to LLM agents remain overwhelmingly API-based: web search, database queries, code execution, and SaaS integrations. While we cannot claim exhaustive coverage of all production systems, our review of major framework documentation (LangChain, AutoGen, CrewAI, n8n) and recent surveys of agentic AI did not surface published examples of pre-trained statistical ML models registered as agent-callable tools.

\subsection{ML-Powered Pricing and Estimation}

Pricing prediction using machine learning is well-established in domains such as real estate, insurance underwriting, and e-commerce dynamic pricing~\citep{ye2018customized}. These systems typically operate as standalone services or batch pipelines, \textit{separate from any conversational AI layer}. The integration of such models within an agentic conversational workflow---where the LLM can reason about and contextualize the prediction---is a key contribution of MLAT.

\subsection{Structured Output Parsing in LLM Systems}

Recent advances in LLM structured output capabilities---including Gemini's JSON schema-constrained generation and OpenAI's structured outputs---have enabled reliable data extraction and inter-agent communication. However, the use of structured outputs to \textit{bridge} LLM reasoning and ML model inputs (extracting features in a schema that maps directly to an ML model's feature vector) has not, to our knowledge, been formalized as a design pattern.

\subsection{Small-Data ML and Synthetic Data Augmentation}

Training ML models with extremely limited data ($N < 100$) is an active research area~\citep{borisov2022deep}. Recent work has explored using LLMs to generate synthetic tabular data that preserves statistical properties of real datasets~\citep{borisov2023language}. MLAT's validation through PitchCraft contributes a case study of combining LLM-generated, human-verified synthetic data with real data for small-scale production ML deployment.

\subsection{Gap in the Literature}

To our knowledge, no prior work has explicitly formalized the pattern of exposing pre-trained ML models as callable tools within LLM agent workflows. While individual practitioners may have implemented similar patterns, the lack of published formalization means there is no shared vocabulary, no documented best practices, and no empirical validation. MLAT aims to fill this gap.

\section{The MLAT Framework}

\subsection{Definition}

\begin{quote}
\textit{Machine Learning as a Tool (MLAT) is a design pattern in which a pre-trained statistical ML model is exposed as a callable function within an LLM agent's tool registry, enabling the agent to invoke the model during its reasoning process, receive the prediction as a tool response, and incorporate it into its output through contextual reasoning.}
\end{quote}

Formally, let $\mathcal{A}$ be an LLM agent with tool registry $\mathcal{T} = \{t_1, t_2, \ldots, t_n\}$. MLAT adds a tool $t_{\text{ml}}$ defined by the following two-step process. Given structured agent context $\mathbf{z}$ (e.g., research findings, transcript data), the agent first extracts a feature vector:

\begin{equation}
    \mathbf{x} = \phi(\mathbf{z}), \quad \mathbf{x} \in \mathbb{R}^d
    \label{eq:features}
\end{equation}

\noindent where $\phi(\cdot)$ is a schema-constrained extraction function (implemented via structured output parsing). The agent then invokes the ML tool:

\begin{equation}
    \hat{y} = t_{\text{ml}}(\mathbf{x}) = f_\theta(\mathbf{x})
    \label{eq:mlat}
\end{equation}

\noindent where $f_\theta$ is the trained model with parameters $\theta$. The prediction $\hat{y}$ is returned to the agent as a tool response for further contextual reasoning.

The agent's decision to invoke $t_{\text{ml}}$ is governed by its own reasoning process, identical to how it decides to invoke web search or database queries. This distinguishes MLAT from static pipelines where ML inference occurs unconditionally.

\subsection{MLAT vs.\ Conventional Integration Patterns}

Table~\ref{tab:patterns} summarizes the key differences.

\begin{table}[H]
\centering
\caption{Comparison of ML integration patterns in AI systems.}
\label{tab:patterns}
\small
\begin{tabularx}{\textwidth}{lXXX}
\toprule
\textbf{Pattern} & \textbf{ML Invocation} & \textbf{LLM Involvement} & \textbf{Contextual Reasoning} \\
\midrule
Static Pipeline & Always, as preprocessing & Post-hoc formatting only & None \\
Ensemble/Hybrid & Parallel with LLM & Combined via voting/weighting & Limited \\
RAG + ML & Retrieval-augmented, ML in retrieval & Generation over retrieved results & Indirect \\
\textbf{MLAT (This Work)} & \textbf{Agent-initiated tool call} & \textbf{Full reasoning loop} & \textbf{Full contextual interpretation} \\
\bottomrule
\end{tabularx}
\end{table}

The fundamental distinction is \textit{agency}: in MLAT, the LLM decides whether, when, and how to invoke the ML model, and can reason about the prediction before incorporating it into the final output.

\subsection{The MLAT Execution Pattern}

Algorithm~\ref{alg:mlat} formalizes the MLAT tool-calling sequence.

\begin{algorithm}[H]
\caption{MLAT Tool-Calling Sequence}
\label{alg:mlat}
\begin{algorithmic}[1]
\State \textbf{Offline Phase:}
\State \hspace{1em} Train model $f_\theta$ on historical data $\mathcal{D}$
\State \hspace{1em} Serialize $f_\theta$; deploy behind REST endpoint $\mathcal{E}$
\State \hspace{1em} Register $\mathcal{E}$ as tool $t_{\text{ml}}$ in agent $\mathcal{A}$'s registry $\mathcal{T}$
\State \hspace{1em} Define extraction schema $\mathcal{S}_{\text{in}}$ and output schema $\mathcal{S}_{\text{out}}$
\Statex
\State \textbf{Runtime Phase:}
\State \hspace{1em} Agent $\mathcal{A}$ receives context $\mathbf{z}$ (transcript, research data)
\State \hspace{1em} $\mathbf{x} \leftarrow \phi(\mathbf{z})$ via schema $\mathcal{S}_{\text{in}}$ \Comment{Feature extraction}
\State \hspace{1em} $\hat{y} \leftarrow f_\theta(\mathbf{x})$ via tool call to $\mathcal{E}$ \Comment{ML prediction}
\State \hspace{1em} Agent $\mathcal{A}$ receives $\hat{y}$ as tool response
\State \hspace{1em} Agent $\mathcal{A}$ reasons about $\hat{y}$ given $\mathbf{z}$ \Comment{Contextual interpretation}
\State \hspace{1em} $\mathcal{O} \leftarrow \mathcal{A}.\text{generate}(\mathbf{z},\, \hat{y},\, \mathcal{S}_{\text{out}})$ \Comment{Structured output}
\end{algorithmic}
\end{algorithm}

\subsection{Design Principles}

MLAT implementations should follow these principles:

\begin{enumerate}[leftmargin=2em]
    \item \textbf{Schema-bridged inputs:} Use structured output schemas to bridge LLM reasoning and ML feature vectors, ensuring reliable feature extraction.
    \item \textbf{Stateless tool calls:} The ML endpoint should be stateless and fast ($<$100ms), matching latency expectations of other agent tools.
    \item \textbf{Prediction transparency:} Return metadata (confidence, feature importances, prediction intervals) to enable richer agent reasoning.
    \item \textbf{Model-agnostic registration:} The tool interface should abstract the underlying model, allowing seamless upgrades without workflow changes.
    \item \textbf{Agent-controlled invocation:} The agent---not the pipeline---decides when to call the ML tool.
\end{enumerate}

\section{Structured Output Architecture with Gemini}

A critical enabler of MLAT is reliable structured output parsing. MLAT requires the LLM agent to extract features in a format that maps directly to the ML model's input schema, and to generate downstream outputs from structured data. Gemini's JSON schema-constrained generation makes this possible at production reliability.

\subsection{Schema-Bridged Feature Extraction}

The Research Agent extracts client intelligence using the following schema, which includes provenance metadata that the downstream Draft Agent uses to assess prediction reliability:

\begin{lstlisting}[language={},caption={Research Agent structured output schema.},label={lst:research_schema}]
{
  "type": "object",
  "properties": {
    "client_revenue": {
      "type": "object",
      "properties": {
        "annual_revenue": { "type": "number" },
        "currency": { "type": "string" },
        "source": { "type": "string" },
        "confidence": {
          "type": "string",
          "enum": ["low", "medium", "high"]
        },
        "year": { "type": "string" }
      },
      "required": ["annual_revenue", "currency",
                    "source", "confidence", "year"]
    },
    "company_summary": { "type": "string" },
    "prospect_summary": { "type": "string" }
  },
  "required": ["client_revenue",
                "company_summary", "prospect_summary"]
}
\end{lstlisting}

The \texttt{annual\_revenue} field maps directly to the ML model's \texttt{client\_revenue} feature, while \texttt{confidence} and \texttt{source} enable the Draft Agent to weight the prediction accordingly.

\subsection{Structured Proposal Generation}

The Draft Agent generates the complete proposal through a detailed JSON schema:

\begin{lstlisting}[language={},caption={Draft Agent structured output schema (abbreviated).},label={lst:draft_schema}]
{
  "type": "object",
  "properties": {
    "project_name": { "type": "string" },
    "primary_goals_intro": { "type": "string" },
    "goals_list": {
      "type": "array", "items": { "type": "string" }
    },
    "deliverables_intro": { "type": "string" },
    "deliverables_list": {
      "type": "array", "items": { "type": "string" }
    },
    "client_requirements": {
      "type": "array", "items": { "type": "string" }
    },
    "timeline_breakdown": {
      "type": "array",
      "items": {
        "type": "object",
        "properties": {
          "week": { "type": "string" },
          "title": { "type": "string" },
          "focus_goal": { "type": "string" },
          "activities": {
            "type": "array",
            "items": { "type": "string" }
          },
          "deliverables": {
            "type": "array",
            "items": { "type": "string" }
          }
        },
        "required": ["week", "title", "focus_goal",
                      "activities", "deliverables"]
      }
    },
    "pricing_section": {
      "type": "object",
      "properties": {
        "total_price": { "type": "number" },
        "currency": { "type": "string" },
        "deposit_amount": { "type": "number" },
        "final_amount": { "type": "number" },
        "value_justification": { "type": "string" }
      },
      "required": ["total_price", "currency",
        "deposit_amount", "final_amount",
        "value_justification"]
    }
  },
  "required": ["project_name", "primary_goals_intro",
    "goals_list", "deliverables_intro",
    "deliverables_list", "client_requirements",
    "timeline_breakdown", "pricing_section"]
}
\end{lstlisting}

The \texttt{pricing\_section.total\_price} is populated using the MLAT prediction. The \texttt{value\_justification} field is where the agent's contextual reasoning materializes---explaining \textit{why} the price is what it is.

\subsection{Schema as Inter-Agent Contract}

The structured output schemas serve as \textbf{contracts between agents}: the Research Agent's output schema guarantees that all fields required by the Draft Agent are present and correctly typed. This is analogous to API contracts in microservice architectures, but applied to LLM-to-LLM communication. Each JSON field in the Draft Agent's output maps to a placeholder in a Google Docs template via the Docs API's Find \& Replace function, enabling fully automated document generation.

\section{PitchCraft: An MLAT Case Study}

To validate the MLAT framework, we built PitchCraft---a pilot production system for automated proposal generation deployed at Legacy AI LLC.

\subsection{Single-Workflow, Dual-Agent Architecture}

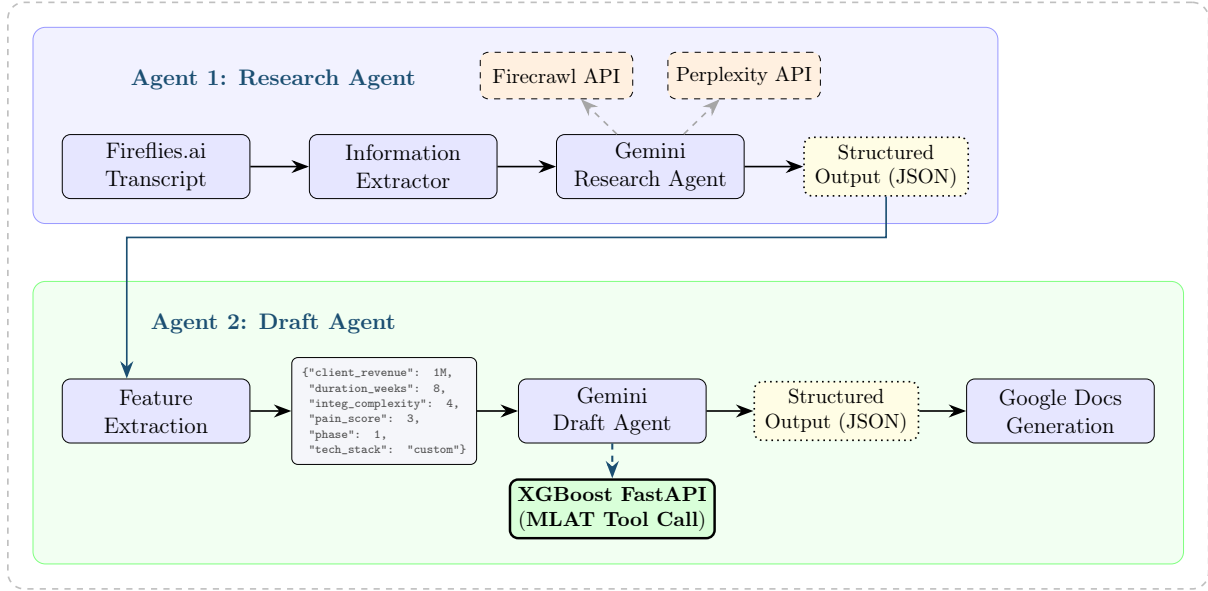
\begin{figure}[H]
\centering
\resizebox{\textwidth}{!}{%
\begin{tikzpicture}[
    node distance=0.7cm and 1.0cm,
    block/.style={rectangle, draw, rounded corners=4pt, fill=blue!10, minimum width=3.2cm, minimum height=1.1cm, align=center, font=\normalsize},
    tool/.style={rectangle, draw, dashed, rounded corners=4pt, fill=orange!12, minimum width=2.6cm, minimum height=0.8cm, align=center, font=\small},
    mlblock/.style={rectangle, draw, line width=1.2pt, rounded corners=4pt, fill=green!15, minimum width=2.8cm, minimum height=0.9cm, align=center, font=\small},
    schema/.style={rectangle, draw, dotted, line width=0.8pt, rounded corners=4pt, fill=yellow!12, minimum width=2.8cm, minimum height=0.9cm, align=center, font=\small},
    jsonbox/.style={rectangle, draw, rounded corners=3pt, fill=codebg, minimum width=2.6cm, align=left, font=\ttfamily\fontsize{6}{7.5}\selectfont, text=black!70, inner sep=5pt},
    arrow/.style={-{Stealth[length=3mm]}, thick},
    label/.style={font=\normalsize\bfseries, text=darkblue}
]

\node[label] (a1label) at (-3.5, 0) {Agent 1: Research Agent};

\node[block] (fireflies) at (-5.5, -1.5) {Fireflies.ai\\Transcript};
\node[block, right=1.0cm of fireflies] (extract) {Information\\Extractor};
\node[block, right=1.0cm of extract] (research) {Gemini\\Research Agent};
\node[schema, right=1.0cm of research] (schema1) {Structured\\Output (JSON)};

\node[tool, above=0.6cm of research, xshift=-1.6cm] (firecrawl) {Firecrawl API};
\node[tool, above=0.6cm of research, xshift=1.6cm] (perplexity) {Perplexity API};

\draw[arrow] (fireflies) -- (extract);
\draw[arrow] (extract) -- (research);
\draw[arrow] (research) -- (schema1);
\draw[arrow, dashed, gray!70] (research) -- (firecrawl);
\draw[arrow, dashed, gray!70] (research) -- (perplexity);

\node[label] (a2label) at (-3.5, -4.2) {Agent 2: Draft Agent};

\node[block] (featext) at (-5.5, -5.7) {Feature\\Extraction};

\node[jsonbox, right=0.7cm of featext] (jsonpayload) {%
\{"client\_revenue": 1M,\\
\ "duration\_weeks": 8,\\
\ "integ\_complexity": 4,\\
\ "pain\_score": 3,\\
\ "phase": 1,\\
\ "tech\_stack": "custom"\}};

\node[block, right=0.7cm of jsonpayload] (draft) {Gemini\\Draft Agent};
\node[schema, right=0.8cm of draft] (schema2) {Structured\\Output (JSON)};
\node[block, right=0.8cm of schema2] (gdocs) {Google Docs\\Generation};

\node[mlblock, below=0.6cm of draft] (xgboost) {\textbf{XGBoost FastAPI}\\(\textbf{MLAT Tool Call})};

\draw[arrow] (featext) -- (jsonpayload);
\draw[arrow] (jsonpayload) -- (draft);
\draw[arrow] (draft) -- (schema2);
\draw[arrow] (schema2) -- (gdocs);
\draw[arrow, dashed, darkblue, line width=1pt] (draft.south) -- (xgboost.north);

\draw[arrow, thick, darkblue] (schema1.south) -- ++(0,-0.7) -| ([xshift=-0.5cm]featext.north);

\begin{scope}[on background layer]
    \node[draw=blue!40, fill=blue!5, rounded corners=6pt,
          fit=(a1label)(fireflies)(extract)(research)(schema1)(firecrawl)(perplexity),
          inner xsep=14pt, inner ysep=12pt] (bg1) {};
    \node[draw=green!50, fill=green!5, rounded corners=6pt,
          fit=(a2label)(featext)(jsonpayload)(draft)(schema2)(gdocs)(xgboost),
          inner xsep=14pt, inner ysep=12pt] (bg2) {};
    \node[draw=gray!50, dashed, line width=0.8pt, rounded corners=8pt,
          fit=(bg1)(bg2),
          inner xsep=12pt, inner ysep=12pt,
          label={[font=\normalsize\bfseries\itshape, text=gray!60]above:Single LLM Workflow}] {};
\end{scope}

\end{tikzpicture}
}
\caption{PitchCraft architecture: a single LLM workflow with two Gemini agents. \textbf{Agent 1} (Research Agent) analyzes the Fireflies transcript and performs parallel tool calls to Firecrawl and Perplexity APIs, outputting structured JSON. \textbf{Agent 2} (Draft Agent) extracts ML features into the model's input schema, invokes the XGBoost pricing model as an \textbf{MLAT tool call} (green, below), reasons about the prediction, and generates the complete proposal via structured output parsing.}
\label{fig:architecture}
\end{figure}

PitchCraft operates as a \textbf{single LLM workflow} with two Gemini-powered agents orchestrated in sequence (Figure~\ref{fig:architecture}).

\subsubsection{Agent 1: Research Agent}

The Research Agent ingests a Fireflies.ai call transcript and performs multi-step intelligence gathering:

\begin{enumerate}[leftmargin=2em]
    \item \textbf{Transcript Ingestion:} A webhook receives Fireflies transcript data. A processing node extracts speaker-segmented text, sentiment analysis, and metadata including questions asked, tasks identified, and pricing mentions.
    \item \textbf{Information Extraction:} A Gemini-powered extractor identifies the prospect name, company name, and project phase.
    \item \textbf{Parallel Tool Calls:} The Research Agent executes tool calls: Firecrawl API for client annual revenue lookup, Perplexity API for company/prospect background research. Pain severity scoring (1--5 scale) and integration complexity scoring (1--5 scale) are performed via dedicated LLM calls within the workflow.
    \item \textbf{Structured Output:} All findings are synthesized into JSON conforming to the research schema (Listing~\ref{lst:research_schema}), including revenue with provenance metadata.
\end{enumerate}

\subsubsection{Agent 2: Draft Agent (MLAT)}

The Draft Agent receives the Research Agent's structured output:

\begin{enumerate}[leftmargin=2em]
    \item \textbf{Feature Extraction:} The agent extracts ML features from the structured research output: \texttt{client\_revenue}, \texttt{estimated\_duration\_weeks}, \texttt{pain\_severity\_score}, \texttt{integration\_complexity}, \texttt{phase}, and \texttt{tech\_stack}.
    \item \textbf{MLAT Tool Call:} The agent invokes the XGBoost FastAPI endpoint as a tool, passing the feature vector (see Figure~\ref{fig:architecture}, JSON payload) and receiving $\hat{y}$.
    \item \textbf{Contextual Reasoning:} The agent reasons about $\hat{y}$ given the research context---adjusting for client relationship, competitive landscape, and strategic factors.
    \item \textbf{Structured Proposal Output:} The agent generates the complete proposal as JSON conforming to the draft schema (Listing~\ref{lst:draft_schema}).
    \item \textbf{Document Generation:} Structured JSON is mapped to a Google Docs template via Find \& Replace.
\end{enumerate}

\subsection{The MLAT Pricing Tool}

\begin{table}[H]
\centering
\caption{Feature schema for the MLAT pricing model.}
\label{tab:features}
\small
\begin{tabularx}{\textwidth}{llXl}
\toprule
\textbf{Feature} & \textbf{Type} & \textbf{Description} & \textbf{Source} \\
\midrule
\texttt{client\_revenue} & Float & Annual revenue of the client & Firecrawl (Agent 1) \\
\texttt{est\_duration\_weeks} & Integer & Projected project timeline & Transcript Analysis \\
\texttt{pain\_severity\_score} & Int (1--5) & Urgency/criticality of need & LLM Scoring \\
\texttt{integration\_complexity} & Int (1--5) & Technical integration difficulty & LLM Scoring \\
\texttt{phase} & Int (1--4) & Project phase number & Info Extraction \\
\texttt{tech\_stack} & Categorical & \texttt{no\_code}/\texttt{low\_code}/\texttt{custom} & Research Agent \\
\bottomrule
\end{tabularx}
\end{table}

The endpoint performs one-hot encoding of \texttt{tech\_stack} at inference time, yielding $\mathbf{x} \in \mathbb{R}^8$. Inference latency is consistently under 100ms.

\section{ML Methodology}

\subsection{Dataset Construction}

The training dataset consists of $N = 70$ records: 40 derived from real agency deals (IDs 2--21, many with multiple project phases) and 30 LLM-generated synthetic records (IDs 22--71). Real data spans 22 industries. Table~\ref{tab:dataset} summarizes the dataset.

\begin{table}[H]
\centering
\caption{Dataset summary statistics ($N = 70$).}
\label{tab:dataset}
\small
\begin{tabular}{lrrrr}
\toprule
\textbf{Feature} & \textbf{Mean} & \textbf{Std} & \textbf{Min} & \textbf{Max} \\
\midrule
\texttt{client\_revenue} (\$) & 8,105,790 & 30,768,920 & 100,000 & 250,000,000 \\
\texttt{est\_duration\_weeks} & 8.4 & 4.5 & 3 & 20 \\
\texttt{pain\_severity\_score} & 3.6 & 0.9 & 2 & 5 \\
\texttt{integration\_complexity} & 3.9 & 0.9 & 2 & 5 \\
\texttt{price} (\$, target) & 16,309 & 11,485 & 2,738 & 40,000 \\
\texttt{phase} & 1.6 & 0.7 & 1 & 4 \\
\bottomrule
\end{tabular}
\end{table}

\begin{figure}[H]
\centering
\includegraphics[width=0.95\textwidth]{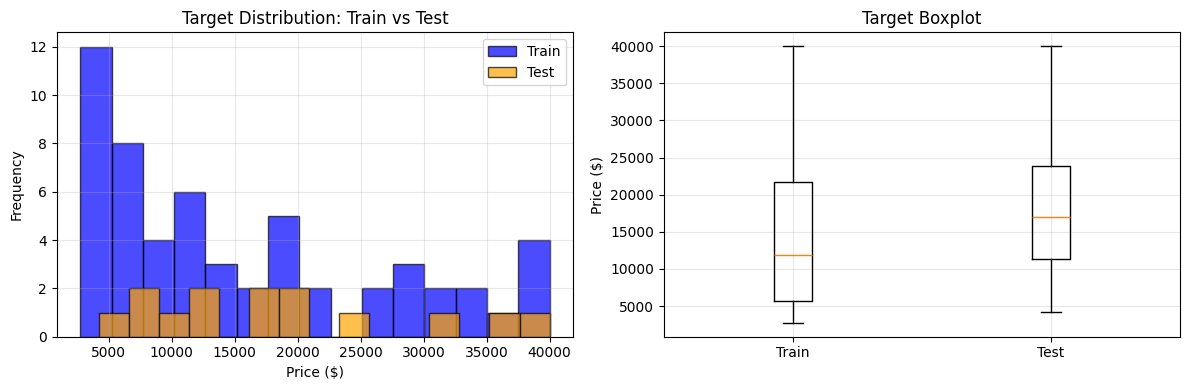}
\caption{Target variable distribution across training and test sets. \textbf{Left:} Histogram showing the right-skewed price distribution. Training (blue) and test (orange) sets show comparable coverage. \textbf{Right:} Box plots confirming similar median and interquartile ranges ($n_{\text{train}}=56$, $n_{\text{test}}=14$).}
\label{fig:distribution}
\end{figure}

The target variable exhibits a right-skewed distribution (Figure~\ref{fig:distribution}). The training and test sets show comparable distributional characteristics, confirming the group-aware splitting strategy did not introduce distributional bias.

\subsection{Synthetic Data Validation}
\label{sec:synthetic}

The 30 synthetic records were generated through a deliberate, human-supervised process:

\begin{enumerate}[leftmargin=2em]
    \item \textbf{Distribution-aware generation:} An LLM was prompted with the statistical distributions (means, standard deviations, correlations, and value ranges) of the 40 real records and instructed to generate new records that extrapolate these distributions to underrepresented industries and price ranges.
    \item \textbf{Human-in-the-loop verification:} All 30 synthetic records were manually reviewed by the authors against real-world pricing knowledge. Records with implausible feature-price relationships (e.g., a high-complexity, long-duration project priced below \$5,000) were edited or regenerated. This manual review ensures the model is not trained on hallucinated noise.
    \item \textbf{Statistical consistency:} Post-generation, we verified that the combined dataset preserved the original data's correlation structure. The synthetic records were assigned unique client group IDs to prevent any data leakage through the group-aware splitting strategy.
\end{enumerate}

This approach is consistent with recent findings that LLM-generated tabular data can preserve multivariate relationships when properly conditioned~\citep{borisov2023language}, while the human verification step provides an additional safeguard against distribution artifacts.

\subsection{Feature Engineering}

With only 70 samples, aggressive feature reduction was critical. We converged on 8 features (7:1 sample-to-feature ratio):

\begin{itemize}[leftmargin=2em]
    \item \textbf{Categorical encoding:} \texttt{tech\_stack} was one-hot encoded into three binary columns, while \texttt{industry} (22 unique values) was excluded to avoid dimensionality explosion.
    \item \textbf{Ablation validation:} Removing \texttt{integration\_complexity} degraded CV $R^2$ from 0.816 to 0.599, confirming it as the most influential feature.
\end{itemize}

\subsection{Data Leakage Prevention}
\label{sec:leakage}

Several real clients had multiple project phases (e.g., avatar: phases 1--3; digital marketing: phases 1--4). We implemented \texttt{GroupShuffleSplit} with manually assigned client groups, ensuring all phases of a multi-phase client were allocated entirely to training or test---never split. This produced a 56/14 train/test split with zero client overlap.

\subsection{Model Selection and Training}

We selected XGBoost~\citep{chen2016xgboost} for strong small-data tabular performance~\citep{borisov2022deep}, built-in regularization, and fast inference. Table~\ref{tab:hyperparams} shows our conservative hyperparameters.

\begin{table}[H]
\centering
\caption{XGBoost hyperparameters.}
\label{tab:hyperparams}
\small
\begin{tabular}{llp{7cm}}
\toprule
\textbf{Hyperparameter} & \textbf{Value} & \textbf{Rationale} \\
\midrule
\texttt{n\_estimators} & 50 & Fewer trees to prevent memorization \\
\texttt{max\_depth} & 3 & Shallow trees for generalization \\
\texttt{learning\_rate} & 0.05 & Slower learning for stability \\
\texttt{subsample} & 0.8 & Row sampling per tree \\
\texttt{colsample\_bytree} & 0.8 & Feature sampling per tree \\
\texttt{reg\_alpha} (L1) & 0.1 & Feature selection via sparsity \\
\texttt{reg\_lambda} (L2) & 1.0 & Weight shrinkage \\
\texttt{min\_child\_weight} & 3 & Minimum 3 samples per leaf \\
\bottomrule
\end{tabular}
\end{table}

\subsection{Cross-Validation Strategy}

Three-fold \texttt{GroupKFold} cross-validation respected client group boundaries:

\begin{table}[H]
\centering
\caption{Cross-validation fold characteristics.}
\label{tab:folds}
\small
\begin{tabular}{lrll}
\toprule
\textbf{Fold} & \textbf{Samples} & \textbf{Price Range} & \textbf{Mean Price} \\
\midrule
Fold 1 & 19 & \$2,738 -- \$40,000 & \$13,952 \\
Fold 2 & 19 & \$4,672 -- \$28,876 & \$16,281 \\
Fold 3 & 18 & \$4,793 -- \$39,022 & \$20,239 \\
\bottomrule
\end{tabular}
\end{table}

\section{Results}

\subsection{Model Performance}

\begin{table}[H]
\centering
\caption{XGBoost pricing model performance.}
\label{tab:performance}
\small
\begin{tabular}{lccc}
\toprule
\textbf{Metric} & \textbf{Training Set} & \textbf{Test Set} & \textbf{Cross-Validation} \\
\midrule
$R^2$ & 0.937 & 0.807 & $0.816 \pm 0.060$ \\
MAE & \$2,328 & \$3,688 & $\$3,898 \pm \$629$ \\
RMSE & \$2,874 & \$4,720 & --- \\
Relative MAE\textsuperscript{a} & 14.3\% & 22.6\% & 23.9\% \\
\bottomrule
\multicolumn{4}{l}{\textsuperscript{a}\scriptsize Relative MAE = MAE / mean target price (\$16,309). Not a true per-sample MAPE.}
\end{tabular}
\end{table}

\begin{figure}[H]
\centering
\includegraphics[width=0.95\textwidth]{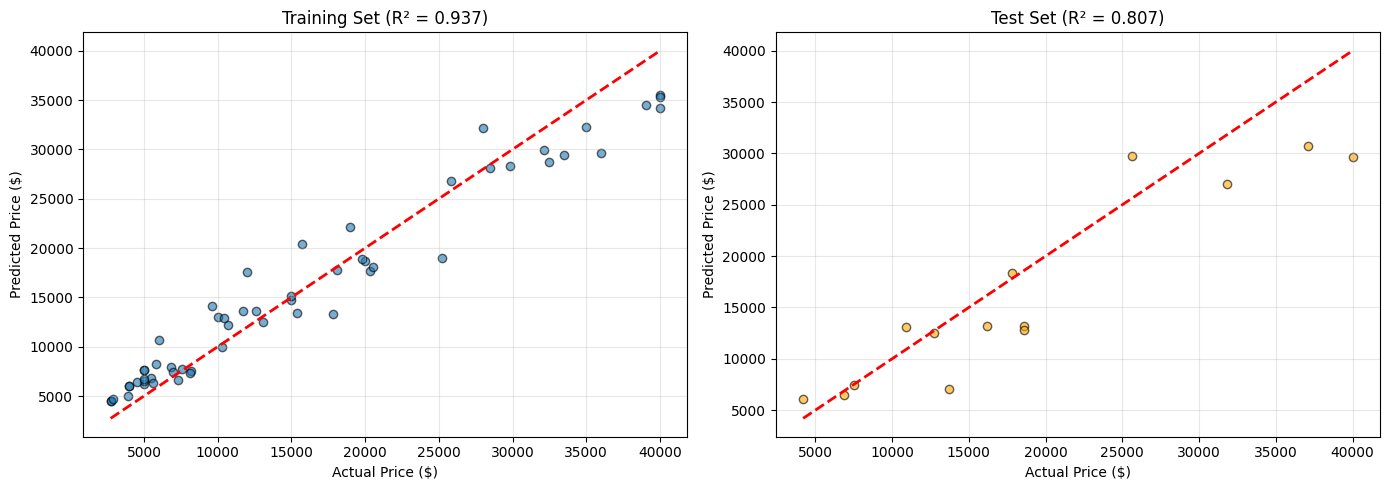}
\caption{Predicted vs.\ actual price. \textbf{Left:} Training set ($R^2 = 0.937$) shows tight clustering around the identity line. \textbf{Right:} Test set ($R^2 = 0.807$) demonstrates generalization to unseen data, with slight overestimation in the mid-range and underestimation at the highest values---consistent with regression-to-the-mean behavior expected with limited tail data.}
\label{fig:predictions}
\end{figure}

The model explains 80.7\% of variance on unseen test data. The close alignment between CV $R^2$ (0.816) and test $R^2$ (0.807) confirms reliable generalization. Figure~\ref{fig:predictions} shows predicted vs.\ actual scatter plots.

\subsection{Model Comparison}

\begin{table}[H]
\centering
\caption{Model comparison: XGBoost vs.\ Ridge regression.}
\label{tab:comparison}
\small
\begin{tabular}{lccc}
\toprule
\textbf{Model} & \textbf{CV $R^2$} & \textbf{CV MAE} & \textbf{Test $R^2$} \\
\midrule
XGBoost (final) & $0.816 \pm 0.060$ & $\$3,898 \pm \$629$ & 0.807 \\
Ridge Regression & $0.565 \pm 0.180$ & $\$4,354 \pm \$249$ & 0.782 \\
\midrule
XGBoost (no \texttt{integ\_complexity}) & $0.599 \pm 0.206$ & $\$4,920 \pm \$815$ & --- \\
\bottomrule
\end{tabular}
\end{table}

XGBoost significantly outperformed Ridge on CV $R^2$ ($0.816$ vs.\ $0.565$), confirming nonlinear feature interactions matter. The ablation removing \texttt{integration\_complexity} degraded performance substantially.

\subsection{Overfitting Analysis}

The train/test RMSE ratio of 0.61 indicates mild overfitting, expected with $N=70$. However, CV-test $R^2$ consistency (0.816 vs.\ 0.807) confirms this is well-controlled.

\subsection{Sensitivity Analysis}
\label{sec:sensitivity}

To verify that the model learned economically meaningful relationships rather than memorizing noise, we performed a univariate sensitivity analysis. We constructed a baseline feature vector representing a ``typical'' project ($\texttt{client\_revenue} = \$1\text{M}$, $\texttt{duration} = 8$ weeks, $\texttt{integration\_complexity} = 3$, $\texttt{phase} = 1$, $\texttt{tech\_stack} = \texttt{custom}$) and varied each feature independently while holding others constant. Table~\ref{tab:sensitivity} shows the results for the two most influential features.

\begin{table}[H]
\centering
\caption{Sensitivity analysis: predicted price (\$) as features vary, holding others at baseline. Monotonic, economically coherent trends confirm the model has learned meaningful relationships.}
\label{tab:sensitivity}
\small
\begin{tabular}{ccc}
\toprule
\textbf{Pain Severity Score} & \textbf{Integration Complexity} & \textbf{Predicted Price (\$)} \\
\midrule
1 & 3 (baseline) & $\sim$\$5,200 \\
2 & 3 & $\sim$\$6,400 \\
3 & 3 & $\sim$\$8,100 \\
4 & 3 & $\sim$\$10,800 \\
5 & 3 & $\sim$\$13,500 \\
\midrule
3 (baseline) & 1 & $\sim$\$4,900 \\
3 & 2 & $\sim$\$6,500 \\
3 & 3 & $\sim$\$8,100 \\
3 & 4 & $\sim$\$11,200 \\
3 & 5 & $\sim$\$14,800 \\
\bottomrule
\end{tabular}
\end{table}

The model produces monotonically increasing prices as pain severity and integration complexity increase, consistent with business logic: more urgent needs and more complex integrations command higher prices. Pain severity 1 vs.\ 5 produces a $\sim$2.6$\times$ price difference; integration complexity 1 vs.\ 5 produces $\sim$3.0$\times$. These are economically plausible multipliers for a professional services pricing model, providing evidence that the model has learned signal rather than noise despite the small dataset.

\subsection{Commercial Viability}

A relative MAE of 22.6\% (computed as MAE / mean target price) is commercially viable for MLAT because the agent doesn't blindly insert the point estimate. The Draft Agent reasons: ``The model predicts \$16,200 based on complexity 4 and an 8-week timeline. Given this client's \$5M revenue and high pain severity, we recommend positioning at \$18,000.'' The ML provides a statistically grounded anchor; the LLM provides the contextual adjustment.

\section{Deployment and Real-World Impact}

\subsection{Pilot Production Deployment}

PitchCraft has been deployed as a pilot production system at Legacy AI LLC, handling proposal generation for the agency's active sales pipeline. We characterize this as ``pilot production'' rather than full production given the model's training on 40 real deals---sufficient for daily use within a single agency's pricing range, but requiring additional data before generalization claims across industries.

\subsection{Measured Impact}

\begin{table}[H]
\centering
\caption{Measured operational impact of PitchCraft deployment.}
\label{tab:impact}
\small
\begin{tabularx}{\textwidth}{lXXl}
\toprule
\textbf{Metric} & \textbf{Before} & \textbf{After} & \textbf{Improvement} \\
\midrule
Proposal creation & 3+ hours & $<$ 10 minutes & 18$\times$ faster \\
Speed-to-lead & 2--3 days & 2--4 hours & 12--18$\times$ faster \\
Pricing consistency & Variable & ML-standardized & Eliminated variance \\
Hands-on time & 3+ hours & $<$ 10 min review & 95\%+ reduction \\
\bottomrule
\end{tabularx}
\end{table}

\subsection{Inference Performance}

The FastAPI-served XGBoost model achieves sub-100ms latency. The complete pipeline completes in 2--3 minutes, dominated by LLM reasoning and document generation.

\section{Discussion}

\subsection{Why MLAT Matters}

MLAT represents a shift in how ML models participate in agentic systems. Rather than passive pipeline components, ML models become \textit{active participants in reasoning loops}. The LLM orchestrates when to invoke the model, interprets its output, and decides how to act. This is powerful for domains where quantitative predictions must be tempered with qualitative judgment---professional services pricing, construction estimation, insurance underwriting.

The structured output architecture (Section~4) is a critical enabler: schema-constrained generation ensures the bridge between LLM reasoning and ML inputs is reliable and type-safe.

\subsection{Generalizability}

MLAT is fundamentally \textbf{industry-agnostic}. The same pattern applies to construction (material costs, labor hours---agent reasons about site conditions), consulting (rates, scope---agent adjusts for relationship factors), insurance (risk, premiums---agent explains rationale), and home services (parts, labor---agent contextualizes urgency). Only the feature engineering and training data change; the MLAT pattern remains constant.

\subsection{Limitations}

\begin{itemize}[leftmargin=2em]
    \item \textbf{Small dataset:} $N=70$ (40 real) limits generalization to novel industries or extreme prices. The sensitivity analysis (Section~\ref{sec:sensitivity}) provides evidence of learned logic, but broader validation is needed.
    \item \textbf{Single-domain validation:} MLAT demonstrated only in agency pricing; multi-domain validation is needed before stronger generalization claims.
    \item \textbf{No feedback loop:} No mechanism for incorporating deal outcomes (won/lost, negotiated price) into model retraining.
    \item \textbf{LLM feature variability:} Pain severity and integration complexity are LLM-assessed, introducing potential input variance across runs.
    \item \textbf{No confidence calibration:} Point estimates without calibrated uncertainty quantification.
    \item \textbf{Synthetic data reliance:} While human-verified (Section~\ref{sec:synthetic}), the 43\% synthetic composition means real-world validation with larger datasets is essential before broader deployment.
\end{itemize}

\subsection{Future Work}

\begin{enumerate}[leftmargin=2em]
    \item Incorporating deal outcomes for expanded training and win/loss prediction as an additional MLAT tool.
    \item Adding prediction intervals for calibrated price ranges.
    \item Multi-domain MLAT deployment (construction, consulting, home services).
    \item Continuous retraining feedback loops.
    \item Evaluating LLM feature extraction stability across model versions.
\end{enumerate}

\section{Conclusion}

We have introduced \textbf{Machine Learning as a Tool (MLAT)}---a design pattern for exposing pre-trained statistical ML models as callable tools within LLM agent workflows. MLAT enables the orchestrating LLM to invoke predictions contextually, reason about them interpretively, and integrate them into structured outputs through schema-constrained generation.

Through PitchCraft, we demonstrated that MLAT is viable in a pilot production setting. Despite extreme data scarcity ($N=70$, 43\% human-verified synthetic), careful feature engineering, group-aware cross-validation, and sensitivity analysis confirmed that the XGBoost pricing model ($R^2 = 0.807$) learned economically meaningful relationships. The structured output architecture using Gemini's JSON schemas enabled reliable inter-agent communication. The system reduces proposal creation from hours to minutes with measurable business impact.

We believe MLAT represents an underexplored but high-value design pattern. As LLM tool-calling matures and multi-agent architectures become standard, integrating statistical ML models as callable tools will combine the structured prediction of classical ML with the contextual reasoning of modern LLMs---a combination more powerful than either alone.

\bibliographystyle{plainnat}

\end{document}